%% file: main_arxiv.tex
\documentclass[letterpaper, 10 pt, conference]{ieeeconf}
\IEEEoverridecommandlockouts
\usepackage{cite}
\usepackage{amsmath,amssymb,amsfonts}
\usepackage{graphicx}
\usepackage{textcomp}
\usepackage{xcolor}
\definecolor{graphteal}{RGB}{0,115,115}
\definecolor{resetbrown}{RGB}{140,70,30}
\usepackage{booktabs}
\usepackage{multirow}

\usepackage{xspace}

\usepackage{algorithm}
\usepackage{algpseudocode}

\usepackage[hidelinks]{hyperref}

\newcommand{\sysname}{HALTER\xspace}
\def\BibTeX{{\rm B\kern-.05em{\sc i\kern-.025em b}\kern-.08em
    T\kern-.1667em\lower.7ex\hbox{E}\kern-.125emX}}
\begin{document}

\title{\LARGE \bf
From Rollout to Reset: A Graph-Based Harness for Autonomous
Long-Horizon Manipulation Evaluation
}

\author{Jing Jiang$^{1,*}$, Yue Yang$^{2,*}$, Xinkai Jiang$^{1}$,
Gedas Bertasius$^{2}$, Daniel J. Szafir$^{2}$, Rudolf Lioutikov$^{1,\dagger}$%
\thanks{$^{*}$These authors contributed equally.
$^{\dagger}$Corresponding author.}%
\thanks{$^{1}$Karlsruhe Institute of Technology, Germany.
{\tt\small unisc@student.kit.edu}, {\tt\small \{xinkai.jiang, lioutikov\}@kit.edu}}%
\thanks{$^{2}$University of North Carolina at Chapel Hill, USA.
{\tt\small \{yygx, gedas, daniel.szafir\}@cs.unc.edu}}%
}

\maketitle
\thispagestyle{empty}
\pagestyle{empty}

\newcommand{\projectpage}{ Project website:
\href{https://yy-gx.github.io/HALTER/}{\underline{link}}.}

\input{section/00_abstract}

\input{section/01_introduction}
\input{section/02_related_work}
\input{section/03_methodology}

\input{section/04_exp}

\input{section/05_conclusion}

\section*{Acknowledgment}

We used Claude and ChatGPT to polish the text of this paper and to generate the
illustrations in Fig.~\ref{fig:teaser} and Fig.~\ref{fig:framework}. The authors
reviewed every generated item, and all technical claims, experiments, and
results are our own.

\bibliographystyle{IEEEtran}
\bibliography{main}

\end{document}

%% file: section/00_abstract.tex
\begin{abstract}
\providecommand{\projectpage}{}
Robot manipulation policies are improving quickly, and real-robot evaluation
remains the standard evidence for that progress. It still relies on a human to
reset the scene between rollouts, which consumes operator time and leaves the
initial state distribution unspecified, so results reproduce poorly. A recent
system, AutoEval, automates both reset and scoring, but only for single-step
tasks, because a long-horizon rollout can terminate in combinatorially many
configurations that no single learned reset policy covers. We present
\sysname, a \underline{H}arness for \underline{A}utonomous
\underline{L}ong-horizon \underline{T}ask \underline{E}valuation and
\underline{R}eset,
which restores the scene by planning over a library of learned atomic reset
skills, so demonstration cost scales with the size of that library rather than
with the number of terminal states.
\sysname builds a spatial scene graph online from point clouds and vision
foundation models, and an LLM reasons over this graph to score the rollout,
plan the reset, and verify that the reset succeeded, without collecting
labeled success images for any task. On four long-horizon tasks on a Franka
arm, \sysname restores the scene in 76\% of episodes, against 52\% for
AutoEval and 65\% for a motion-planning reset, and it estimates the
completed-skill fraction correctly in 90\% of episodes, against 76\%. 
Its reset-verification verdict is correct in 91\% of episodes, compared with 78\% for AutoEval. It also cuts the operator time of an evaluation campaign by
72\% relative to manual reset. We further measure compositional generalization on three
held-out tasks, where \sysname resets 74.7\% of episodes against 1.3\% for a
per-task reset policy, and we ablate the scene representation and the graph
update rate.\projectpage
\end{abstract}

%% file: section/01_introduction.tex
\section{Introduction}
\label{sec:intro}

\begin{figure}[t]
    \centering
    \includegraphics[width=\linewidth]{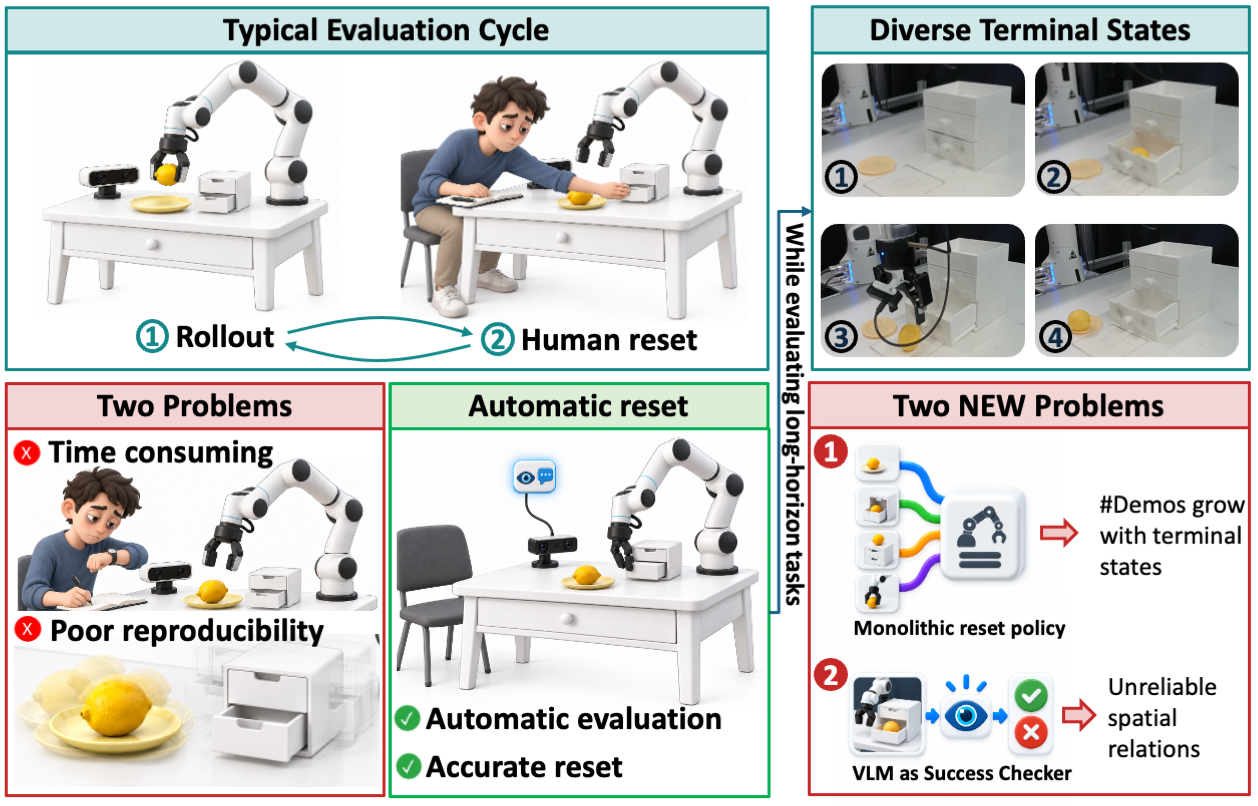}
    \caption{Real-robot evaluation alternates a rollout with a reset (left).
    A human reset costs operator time and leaves the initial state unspecified,
    and an automated reset removes the human for single-step tasks. A
    long-horizon rollout, however, stops in many different terminal states
    (right): this four-step task fails at opening the drawer (1), at removing
    the lemon (2), in transit (3), and at closing the drawer (4). Covering them
    breaks both parts of an automated reset, because a monolithic reset policy
    needs demonstrations that grow with the number of terminal states, and a
    single-image success checker reads spatial relations unreliably.}
    \label{fig:teaser}
\end{figure}

Robot learning algorithms have advanced rapidly in recent years, and
manipulation policies now handle an increasingly broad range of tasks~\cite{intelligence2026pi, kim2025fine, kim2026cosmos, zhang2026native}. As the
number of such policies grows, fair and efficient evaluation becomes the
foundation on which the field measures progress. Simulation already
supports this. A simulator restores the scene to a common initial distribution
before every trial and executes many rollouts in parallel, which makes
evaluation both reproducible and inexpensive~\cite{mittal2025isaac, tao2024maniskill3, zhu2020robosuite}. Real-robot evaluation
lags well behind. The operator restores the initial state according to
individual judgment, so the initial state distribution remains unspecified and
results reproduce poorly across trials and across laboratories~\cite{khargonkar2024scenereplica, barreiros2026careful}.
Restoring every object by hand also consumes substantial operator time, which
bounds how many rollouts a campaign can afford
(Fig.~\ref{fig:teaser}, left). Even recent real-robot
benchmarks still depend on manual reset, and they identify it as a
non-parallelizable bottleneck on evaluation scale~\cite{chen2026robodojo, jangir2025robotarena}.

AutoEval~\cite{zhou2025autoeval} removes the human
operator from this loop. It learns a task-specific reset policy and success
classifier, which lets it evaluate policies continuously without supervision.
This design assumes that a rollout ends in a narrow set of configurations,
which holds for the single-step tasks it targets. Long-horizon tasks violate
that assumption. A policy may fail at any skill, and a single task often
admits many valid skill orderings, so the set of reachable terminal states
grows combinatorially with the length of the task
(Fig.~\ref{fig:teaser}, right). Learning one reset policy
that covers all of them is impractical, because it would require a
combinatorial number of demonstrations~\cite{yang2026lilo}. We therefore ask: \textit{how can a
system automatically reset the scene for long-horizon manipulation tasks in
real-world robot experiments?}

We present \sysname, a system that evaluates long-horizon manipulation
policies on real hardware with little human intervention. \sysname removes this
blocker with a hierarchical reset, in which a high-level planner sequences
low-level atomic skills to restore the scene. Demonstration cost then scales
with the size of the skill library rather than with the number of terminal
states, because the planner recombines a fixed set of skills instead of
learning a separate policy for every state. Such a design depends on
understanding the scene state, which \sysname needs in three places: scoring
the rollout, planning the reset, and verifying that the reset succeeded.

Recent systems commonly ask a vision-language model to read the scene state
directly from images~\cite{zhou2025autoeval, grislain2025failsense}. An image, however, leaves object geometry implicit in its
pixels, and current VLMs read the fine spatial
relations~\cite{chen2024spatialvlm, pothiraj2025capture} that all three steps
require unreliably. \sysname
therefore separates perception from reasoning. It recovers geometry from point
clouds and vision foundation models~\cite{oquab2023dinov2, ravi2025sam}, assembles the result into a
spatial graph over objects and their relations, and leaves the reasoning over
that graph to an LLM, so each component operates where it is strong. \sysname
builds this graph online as the rollout proceeds. The scene is only partially
observable from any single view, so temporal information is essential. Online
construction aggregates observations across the horizon, which lets the graph
retain state that is observed at one moment and occluded later. On four long-horizon tasks on a Franka arm, \sysname
resets the scene more reliably than both a per-task reset policy and a
motion-planning reset, and it extends to held-out tasks without new
demonstrations.

We make the following contributions:
\begin{itemize}
    \item We present the first framework for automatic evaluation of
    long-horizon manipulation tasks on real robots, which restores the scene
    with a hierarchical planner over atomic reset skills.
    \item We propose a graph-based scene state estimator that pairs point
    clouds and vision foundation models with LLM reasoning, and requires no
    per-task trained classifier.
    \item We evaluate \sysname on a Franka arm across four long-horizon
    tasks, where it restores the scene in 76\% of episodes against 52\% for a
    per-task reset policy and 65\% for a motion-planning reset. On three held-out
    tasks the planner recombines the same skill library into new reset
    sequences, so a held-out task costs no new demonstrations, and an ablation
    shows that the scene graph accounts for the gain.
\end{itemize}

%% file: section/02_related_work.tex
\section{Related Work}
\label{sec:related}

\subsection{Real-Robot Policy Evaluation}

Early real-robot benchmarks focused on reproducibility.
REPLAB~\cite{yang2019replabreproduciblelowcostarm} standardizes the hardware and
fixes a protocol for scattering the objects, and
SceneReplica~\cite{khargonkar2024scenereplica} generates scenes in simulation and
has an operator rebuild them on the table.
FurnitureBench~\cite{heo2023furniturebenchreproduciblerealworldbenchmark} carries
the idea to long-horizon assembly. These benchmarks constrain the initial state,
but a person still produces it before every rollout, and rigorous protocols still
rely on human raters to judge the outcome~\cite{barreiros2026careful}.

A more recent line targets the operator time that these protocols require.
RoboDojo~\cite{chen2026robodojo} centralizes the hardware and opens it to remote
submission, while
RoboArena~\cite{atreya2025roboarenadistributedrealworldevaluation} distributes
evaluation across institutions and raters. Both keep the human in the loop, so
they raise throughput without lowering the cost of a single rollout.
AutoEval~\cite{zhou2025autoeval} removes the operator instead, learning a reset
policy and a success classifier for each task, which its single-step setting
makes tractable. Concurrent systems also reset real scenes without a
human, for data collection and for policy
self-improvement~\cite{xiao2026enpireagenticrobotpolicy,
wang2026radarclosedlooproboticdata, wang2026zero2skillbootstrappingrobotskills},
though each prepares the reset for a task fixed in advance, by inverting the
plan it just executed or by synthesizing a routine for that task. \sysname
instead plans the reset at run time from the estimated terminal state of an
external policy that may fail at any subgoal. The word harness also carries a
different sense in recent work: Harness VLA~\cite{zhang2026harnessvlasteeringfrozen}
and Guava~\cite{liu2026guavaeffectiveuniversalharness} wrap a manipulation
policy in order to steer it, whereas \sysname wraps one in order to measure it.

\subsection{Long-Horizon Manipulation}

Long-horizon manipulation is commonly solved by composition. Task and motion
planning sequences predefined skills under a symbolic
planner~\cite{kaelbling2011hierarchical, garrett2021integrated}. Learned variants
replace those skills with policies: LEAGUE~\cite{cheng2023league} lets a symbolic
planner guide skill learning, GSC~\cite{mishra2023generative} chains skills with
diffusion models, and
NOD-TAMP~\cite{cheng2024nodtampgeneralizablelonghorizonplanning} transfers them
through neural object descriptors. Plan-Seq-Learn~\cite{dalal2024plan} and
ManipGen~\cite{dalal2025local} pair a language-model planner with learned local
policies. Scene graphs serve such planners as the state representation:
Zhu et al.~\cite{zhu2021hierarchical} plan long-horizon manipulation over
geometric and symbolic scene graphs, MomaGraph~\cite{ju2026momagraph} builds
state-aware scene graphs with vision-language models for embodied task
planning, and Yu et al.~\cite{yu2025scene} read a scene graph to detect a
failure and replan around it. \sysname reuses this machinery and claims only
the reset system that it supports. Composition is attractive for one
reason. The number of distinct configurations a task can reach grows with its
horizon, so any method that needs data or code for each configuration stops
scaling, while a fixed skill library recombines into many of them.

Recent policies extend the horizon inside a single model instead.
Long-VLA~\cite{fan2025long} and RoboTTT~\cite{jiang2026robotttcontextscalingrobot}
scale a single policy to tasks with many stages, and they meet the same growth.
LiLo-VLA~\cite{yang2026lilo} and BOSS~\cite{yang2025boss} show that such a
rollout can break at any subgoal, so the configurations a policy leaves behind
multiply with the number of subgoals the task contains.
Foresight~\cite{zhang2026foresightfailuredetectionlonghorizon} detects failures
from rollout history, and REFLECT~\cite{liu2023reflect} explains them and prompts
a planner to correct the task. Neither restores the scene for the next rollout.
\sysname composes skills in the same way, but to undo the rollout rather than to
perform it.

%% file: section/03_methodology.tex
\section{Methodology}
\label{sec:methodology}

We present \sysname in four parts. Section~\ref{sec:formulation} formalizes the
task, the rollout, and the reset problem, and it identifies the three questions
that an autonomous evaluation system must answer. Section~\ref{sec:hierarchy}
presents the hierarchy \sysname uses to reset the scene.
Section~\ref{sec:scene_graph} describes the scene representation, and
Section~\ref{sec:graph_uses} describes how \sysname answers the three questions
from that representation.

\begin{figure*}[t]
    \centering
    \includegraphics[width=\textwidth]{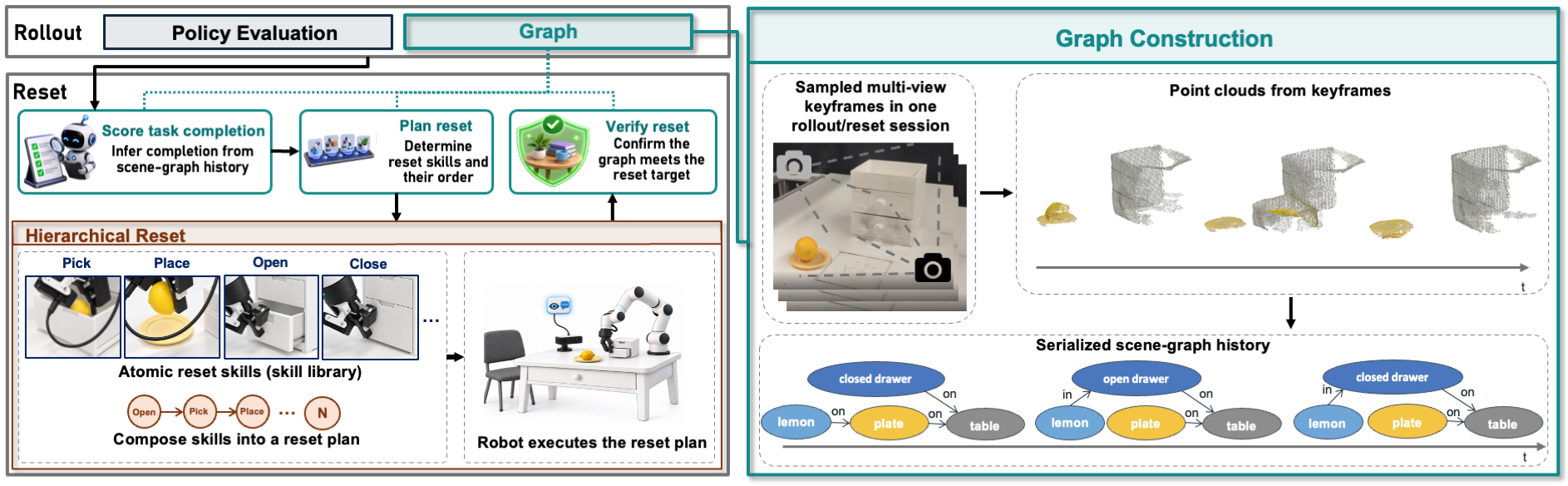}
    \caption{\textbf{Overview of \sysname.} During a rollout, \sysname samples
    multi-view RGB-D keyframes, recovers a point cloud for every object, and
    serializes the result into a scene-graph history (right). A shared LLM
    interface reads that history to score task completion, to plan the reset,
    and to verify it (top left), and the planner composes atomic skills from
    the library into the plan that the robot executes (bottom left). \textbf{Black}
    arrows give the execution order, \textbf{\textcolor{graphteal}{green}} marks
    the graph construction, and \textbf{\textcolor{resetbrown}{brown}} marks the hierarchical
    reset.}
    \label{fig:framework}
\end{figure*}

\subsection{Problem Formulation}
\label{sec:formulation}

A natural language instruction $x$ specifies a long-horizon manipulation task.
Following the task and motion planning (TAMP)
formulation~\cite{garrett2021integrated}, we write the task as a set of atomic
skills $\mathcal{S} = \{a_1, \ldots, a_H\}$ together with a causal partial order
$\prec$ that their preconditions and effects induce, where each skill $a_h$
applies an operator such as \texttt{Pick} or \texttt{Open} to a reference object
$o_h$. Any total order consistent with $\prec$ executes the task correctly, and
we write $\mathrm{Lin}(\mathcal{S}, \prec)$ for the set of these linearizations.
At each step a policy $\pi$ under evaluation observes an RGB image $I_t$ and the
proprioceptive state $q_t$, and it emits an action $u_t$. A rollout starts from
a physical scene state $s_0$ and ends at a terminal state $s_T$.

Evaluating a policy requires repeated rollouts from the same initial condition.
We specify that condition by relations between objects rather than by object
poses, $\mathcal{C}_0 = \{ r_j(o_j, o_j') \}$, for example
$\{\texttt{on(lemon,plate)}, \texttt{on(plate,table)}\}$. Such a target
constrains the relations and leaves each object pose free to vary within them.
We state the problem this work addresses as follows. \emph{Reset problem:}
given the terminal state $s_T$ of a rollout and a target $\mathcal{C}_0$, find
a sequence of robot actions that carries the scene from $s_T$ to a state $s$
satisfying $s \models \mathcal{C}_0$.

Repeating this process autonomously raises two subproblems.
\emph{(P1) Terminal state coverage.} A policy under evaluation can fail at any
skill, and it can follow any of the linearizations in
$\mathrm{Lin}(\mathcal{S}, \prec)$, so the number of terminal states a rollout
can reach grows combinatorially with the task length $H$. Any
reset that maps a terminal state directly to actions has to cover that set, and
the demonstrations it needs grow with it. \emph{(P2) Scene state estimation.}
Meanwhile, the system has to read the state of the scene at three points:
(A) which skill effects held during the rollout, which gives a completion score
$S_{\mathrm{comp}} \in [0,1]$, the fraction of the $H$ skills the rollout
completed; (B) what the robot must do to bring the scene to a state satisfying
$\mathcal{C}_0$; and (C) whether the state the robot reaches satisfies
$\mathcal{C}_0$. All three read the same object: (A) and (C) evaluate relations
on the scene state, and (B) plans over it.

\subsection{Hierarchical Reset}
\label{sec:hierarchy}

Subproblem (P1) shows that any reset which maps a terminal state directly
to actions needs data that grows with the number of reachable terminal states.
\sysname therefore separates the reset into two layers.
Fig.~\ref{fig:framework} (bottom left) shows the resulting two layers. At the low level, instead
of preparing a complete reset policy for each terminal state, we prepare a set
of atomic skills $\mathcal{K} = \{k^{(1)}, \ldots, k^{(M)}\}$, where each skill
performs one thing, such as picking up an object or closing a drawer (we assume
this set suffices to return any reachable terminal state to a state satisfying
$\mathcal{C}_0$). This library is its own set and need not coincide with the
skills $\mathcal{S}$ that the evaluated task performs. The amount of data then depends on the size of the skill set
rather than on the number of terminal states. We learn these skills with a VLA,
because one VLA architecture covers diverse skills that act on different kinds
of objects, from rigid objects to containers to articulated objects. At the high
level, a planner reads the current scene state and the target $\mathcal{C}_0$,
decides which skills to invoke and in what order, and emits an ordered reset
plan $P = (k_1(\alpha_1), \ldots, k_N(\alpha_N))$. The same set of skills
therefore composes into different reset sequences for different terminal states.

\subsection{Scene Graph Construction}
\label{sec:scene_graph}

Subproblem (P2) requires the system to read the state of the scene at three
points. A raw image leaves object geometry implicit in its pixels, and a raw
point cloud carries geometry without grounding it on named objects, so neither
supports the relation tests that (A) and (C) perform nor the search space that
(B) needs. \sysname therefore represents the scene as a spatial
graph~\cite{zhu2021hierarchical, ju2026momagraph, yu2025scene}, which states
objects, their metric attributes, and the relations between
them explicitly, so that (A) and (C) evaluate relations directly on the graph
and (B) grounds every step of its plan on a named object. \sysname builds one
graph from each sampled RGB-D frame, as Fig.~\ref{fig:framework} (right) shows. Given the names of the task-relevant objects,
it segments every object instance in the RGB image with an open-vocabulary
segmentation model~\cite{ren2024grounded, liu2024grounding}, projects the resulting
masks into the registered depth image, and recovers a point cloud for each
object. A node carries an object label together with geometric attributes that
\sysname derives from that point cloud, such as position and spatial extent.
For the edges, we design a set of geometry-based functions that take a pair of
object point clouds and annotate the pair with a spatial relation such as
\texttt{on}, \texttt{in}, or \texttt{under}. These functions read geometric
quantities alone, so the same scene always yields the same annotation.

The moment at which \sysname builds these graphs matters as much as their
content. The environment is partially observable by nature: the state inside a
container or a drawer does not become visible by adding cameras or changing
viewpoints. Observing the scene only after a rollout ends therefore cannot
recover the state it stopped in. For example, no camera placement reveals
whether the lemon sits inside a closed drawer. Reset has an advantage here,
because it always follows a rollout that the system could observe from
beginning to end, and the information that resolves such ambiguity is visible
at the moment the state changes. This history is available to any evaluation
system, yet only a system that accumulates state while the rollout runs retains
it. \sysname therefore builds a graph at a fixed sampling rate during the
rollout instead of estimating the state after the rollout ends. When an object
becomes occluded, \sysname keeps its most recent valid observation together
with the relations it supports, and updates both when the object reappears. The
resulting sequence of graphs is the common input to the three uses in
Section~\ref{sec:graph_uses}.

\subsection{Scoring, Planning, and Verification}
\label{sec:graph_uses}

\sysname uses an LLM for all three reasoning steps
(Fig.~\ref{fig:framework}, top left). It serializes each graph
into object records and relation triples, so the three steps read the same
input and differ only in their prompt.

Evaluation first has to report what the rollout achieved. The LLM reads the
instruction and the graph sequence, decides for each skill whether its effect
held at some point during the rollout, and returns
$\widehat{S}_{\mathrm{comp}}$. The graph sequence matters here, because the
effect of a skill can hold in the middle of a rollout and then become occluded
or undone, and a reading taken at the terminal state alone misses it.

Once the rollout is scored, the scene has to return to $\mathcal{C}_0$, and the
hierarchy of Section~\ref{sec:hierarchy} assigns that decision to a planner
above the skill library. The LLM reads the current state from the graph
sequence, together with $\mathcal{C}_0$, the skill library $\mathcal{K}$, and a
set of reference cases, and it returns the ordered reset plan $P$. Every skill
argument in the plan names a node that the graph already contains, so the plan
is a sequence the robot can execute rather than a description. The LLM chooses
which skills to call and in what order, and the VLA skills produce the actions.

Executing a plan does not guarantee that the scene reached $\mathcal{C}_0$.
\sysname therefore verifies at two levels. After each skill it checks whether
the expected effect of that skill appears in the new graph. A missing effect
stops the plan, because the remaining skills assume that this one succeeded.
When the plan runs to completion, \sysname checks the final graph against
$\mathcal{C}_0$, and it requests human intervention if either check fails.
Without this verification, the next rollout could begin from an unverified
state. Checking the graph stands as a proxy for checking the physical scene.

%% file: section/04_exp.tex
\section{Experiments}
\label{sec:experiments}

Our experiments ask three questions. \textbf{(Q1)} Does \sysname make
autonomous evaluation of long-horizon tasks work with little human
involvement? \textbf{(Q2)} Does the reset hierarchy transfer to held-out tasks
without new demonstrations? \textbf{(Q3)} Does the scene graph of
Section~\ref{sec:scene_graph} account for these results? Section~\ref{sec:exp_setup} and Section~\ref{sec:impl}
describe the setup, and the three subsections that follow answer the questions
in order.

\subsection{Experimental Setup}
\label{sec:exp_setup}

\subsubsection{Tasks}
\begin{figure*}[t]
    \centering
    \includegraphics[width=\textwidth]{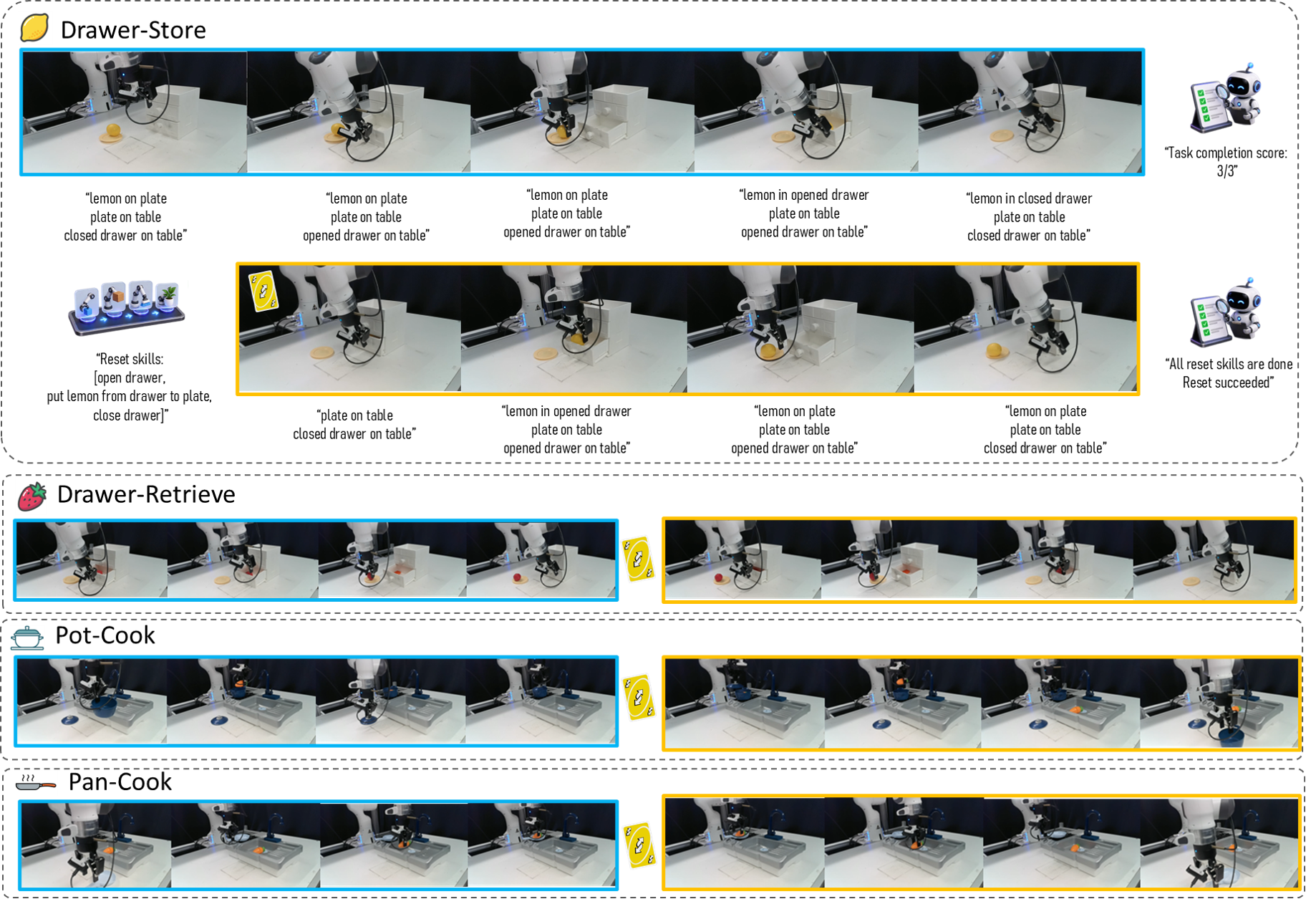}
    \caption{Four evaluated long-horizon tasks across two tabletop scenes.}
    \label{fig:tasks}
\end{figure*}

We evaluate on four long-horizon tasks in two tabletop scenes, which
Fig.~\ref{fig:tasks} shows. In the drawer scene, \emph{Drawer-Store} opens the
lower drawer, places a lemon inside, and closes it, and \emph{Drawer-Retrieve}
opens the drawer, takes a strawberry out, and places it on a plate. In the
stove scene, \emph{Pot-Cook} places a stock pot on the stove and cooks a carrot
in it, and \emph{Pan-Cook} places a pan on the stove and cooks a carrot in it.
Every task takes several dependent steps, so a rollout can stop in many
partially completed states, and each of them calls for a different reset
sequence. For each of the four long-horizon tasks and three held-out tasks, all compared reset methods evaluate rollouts generated by the
same task-specific BESO~\cite{reuss2023goal} policy checkpoint. Each rollout terminates when it
completes the task or reaches a fixed time limit. We run 25 episodes for every task and method.

\subsubsection{Real-Robot Setup}
All experiments run on a Franka Panda 7-DoF arm with a Robotiq 2F-85 parallel
gripper. Three RGB-D cameras observe the workspace: two DepthAI OAK-D Lite
cameras, one at the side of the table and one on the gripper, and one ZED
camera on the opposite side.

\subsubsection{Baselines}
We compare \sysname against two systems.
\textbf{AutoEval}~\cite{zhou2025autoeval} learns one task-specific reset policy
per task. To hold the LLM backbone constant, we use the same LLM as \sysname for
task scoring and post-reset verification. Its reset policy is an
X-VLA~\cite{zheng2025xvla} fine-tuned on complete reset trajectories, with 30 to
40 demonstrations for each failure case and for the successful case.
\textbf{MP skills} keeps every component of \sysname except the skills
themselves, and executes each planned skill with a motion-planning stack:
GraspNet generates the grasp pose and CuRobo plans the motion that reaches it.

\subsubsection{Metrics}
\textbf{Scoring accuracy} is the fraction of episodes whose predicted
completion score $\widehat{S}_{\mathrm{comp}}$ matches the manually annotated
completed-skill fraction.
\textbf{Plan validity} is the fraction of episodes with a valid reset-skill
sequence for restoring $\mathcal{C}_0$; it does not apply to AutoEval, which
emits actions rather than plans.
\textbf{Verification accuracy} is the fraction of episodes in which the
post-reset verification verdict matches the manual annotation of whether
$\mathcal{C}_0$ was restored.
\textbf{Reset SR} is the fraction of episodes whose reset restores
$\mathcal{C}_0$, according to the manual final-scene annotation.
All four metrics use 25 episodes per task, and the main-task averages pool 100
episodes. Uncertainties are binomial standard errors over episodes. The last three
metrics measure cost. \textbf{Interventions} counts episodes requiring a human,
\textbf{cycle time} is the wall-clock seconds per episode, and
\textbf{operator time} charges every intervention the cost of one manual reset.

\subsection{Implementation Details}
\label{sec:impl}

\subsubsection{Reset Skill Policies}

For each atomic skill in $\mathcal{K}$, we train two X-VLA policies using
30 to 50 demonstrations per skill. The drawer-related tasks require six skills:
\emph{Open the lower drawer}, \emph{Close the lower drawer},
\emph{Put the lemon from the drawer back on the plate},
\emph{Put the lemon from the table back on the plate},
\emph{Put the strawberry from the plate back in the drawer}, and
\emph{Put the strawberry from the table back in the drawer}.

The kitchen-related tasks likewise require six skills:
\emph{Put the pot lid back in place},
\emph{Put the stock pot from the stove back in place},
\emph{Put the stock pot from the table back in place},
\emph{Put the pan from the stove back in place},
\emph{Put the pan from the table back in place}, and
\emph{Put the carrot back in the sink}.

\subsubsection{Scene Graph Construction}
\sysname samples RGB-D keyframes at 0.5\,Hz, segments task-relevant objects
using GroundedSAM~\cite{ren2024grounded, liu2024grounding}, and projects the
resulting masks into the registered depth image to recover object-level point
clouds. It infers spatial relations from world-coordinate bounding boxes:
\texttt{on} from vertical proximity and horizontal support, \texttt{in} from
volumetric containment. Drawer states are determined from their
positions along the world $y$-axis.

\subsubsection{LLM Prompting}

We use Qwen3.5-397B for task-completion scoring, reset planning, and reset verification, with a stage-specific prompt for each function. For each task, the scoring and planning prompts contain four task-specific rollout reference cases: one successful execution and three failed executions. Each case pairs a chronological scene-graph history with the corresponding reset-skill sequence. Reset verification uses four successful, skill-specific reference cases. The rollout cases are defined separately for each task, whereas reset cases can be reused across tasks that invoke the same atomic skill.

\begin{table*}[t]
  \centering
  \caption{Main results on the four long-horizon tasks.}
  \label{tab:main}
  \scriptsize
  \setlength{\tabcolsep}{6pt}
  \renewcommand{\arraystretch}{1.15}
  \begin{tabular}{llrrrr@{\hspace{14pt}}c}
    \toprule
    & & \multicolumn{4}{c}{Task (\%)} & \\
    \cmidrule(lr){3-6}
    Method & Metric & Drawer-Store & Drawer-Retrieve & Pot-Cook & Pan-Cook & Avg (\%) \\
    \midrule
    \multirow{4}{*}{AutoEval}
      & Scoring accuracy      & 72 & 64 & 80 & 88 & $76.0 \pm 4.3$ \\
      & Plan validity         & -  & -  & -  & -  & - \\
      & Verification accuracy & 76 & 80 & 80 & 76 & $78.0 \pm 4.1$ \\
      & Reset SR              & 52 & 64 & 16 & 76 & $52.0 \pm 5.0$ \\
    \midrule
    MP skills
      & Reset SR              & 68 & 76 & 72 & 44 & $65.0 \pm 4.8$ \\
    \midrule
    \multirow{4}{*}{\sysname (ours)}
      & Scoring accuracy      & 100 & 96 & 80 & 84 & $\mathbf{90.0 \pm 3.0}$ \\
      & Plan validity         & 92  & 84 & 72 & 88 & $84.0 \pm 3.7$ \\
      & Verification accuracy & 96 & 84 & 88 & 96 & $\mathbf{91.0 \pm 2.9}$ \\
      & Reset SR              & 88  & 84 & 48 & 84 & $\mathbf{76.0 \pm 4.3}$ \\
    \bottomrule
  \end{tabular}
  \vspace{2.5\baselineskip}
\end{table*}

\begin{table}[tb]
  \centering
  \caption{Human effort and time over the 100 episodes. Operator time charges
  each intervention one manual reset.}
  \label{tab:cost}
  \scriptsize
  \setlength{\tabcolsep}{5pt}
  \renewcommand{\arraystretch}{1.15}
  \begin{tabular}{lccc}
    \toprule
    Method & Interventions & Cycle time (s/ep) & Operator time (min) \\
    \midrule
    Manual reset    & 100         & 53  & 88 \\
    AutoEval        & 46          & 104 & 44 \\
    MP skills & 34          & 137 & 29 \\
    \sysname (ours) & \textbf{25} & 121 & \textbf{24} \\
    \bottomrule
  \end{tabular}
\end{table}

\subsection{Autonomous Evaluation and Reset (Q1)}
\label{sec:q1}

Table~\ref{tab:main} reports the main results. \sysname restores the scene in
76.0\% of the episodes, which is 24 points above AutoEval and 11 points above
MP skills. Pooled over the 100 episodes, the gap over AutoEval is more than
three standard errors. AutoEval trains one reset policy on demonstrations that span a whole
long-horizon reset, and a single policy degrades as that horizon grows, whereas
every skill of \sysname covers one step. MP skills shares the scoring and the
planning of \sysname, so the table compares it on end-to-end reset alone: it
picks a grasp pose once and then executes a fixed trajectory, so a small
deviation is enough to fail the skill, whereas the learned skills act on the
current observation. Scoring accuracy reaches 90.0\% and plan validity 84.0\%,
so each step of the graph pipeline is reliable on its own and not only in
combination.

Verification separates the two systems further. Across the 100 episodes,
\sysname achieves a verification accuracy of 91.0\%, against 78.0\% for
AutoEval. These errors have
asymmetric costs: false acceptance can start the next rollout from an incorrect
state, whereas false rejection triggers an unnecessary reset attempt or human
intervention.

Table~\ref{tab:cost} reports what the evaluation costs a human. \sysname cuts
the interventions from 100 to 25 and the operator time from 88 to 24 minutes,
and it raises the mean number of episodes between interventions from 1.0 to 4.0.
Additional LLM reasoning between successive reset skill executions makes each episode slightly longer in wall-clock time than for AutoEval. The operator, however, takes part in far fewer
episodes, and operator availability rather than wall-clock time bounds how many
rollouts an evaluation campaign can afford. The advantage also grows with the
length of the reset plan. Stratified by that length on the three tasks for which
we measured it, \sysname finishes a one-skill reset 2 to 18 seconds faster than
MP skills, and a three-skill reset 24 to 57 seconds faster, with every
three-skill gap significant at $p < 0.01$. Longer resets are exactly the case
that a long-horizon rollout produces.

Pot-Cook is the weakest task for every method, and it is the only task on which
MP skills resets more reliably than \sysname, 72\% against 48\%. It carries the
longest reset plans of the four tasks, and the learned skills handle the stock
pot less reliably than the analytic grasps do, so the gap over MP skills rests
on the other three tasks. \sysname still resets Pot-Cook three times as often as
AutoEval, 48\% against 16\%, and it scores the rollout correctly on 80\% of the
episodes there.

\begin{table}[tb]
  \centering
  \caption{Transfer to three held-out tasks, 25 episodes each.
  \sysname reuses its skill library and trains nothing new.}
  \label{tab:compositional}
  \scriptsize
  \setlength{\tabcolsep}{4pt}
  \renewcommand{\arraystretch}{1.15}
  \begin{tabular}{llrrr@{\hspace{10pt}}c}
    \toprule
    & & \multicolumn{3}{c}{Held-out task (\%)} & \\
    \cmidrule(lr){3-5}
    Method & Metric & Store & Retrieve & Pot & Avg (\%) \\
    \midrule
    \multirow{4}{*}{AutoEval}
      & Scoring accuracy      & 72 & 72 & 80 & $74.7 \pm 5.0$ \\
      & Plan validity         & -  & -  & -  & - \\
      & Verification accuracy & 84 & 80 & 84 & $82.7 \pm 4.4$ \\
      & Reset SR              & 0  & 4  & 0  & $1.3 \pm 1.3$ \\
    \midrule
    \multirow{4}{*}{\sysname (ours)}
      & Scoring accuracy      & 100 & 100 & 84 & $\mathbf{94.7 \pm 2.6}$ \\
      & Plan validity         & 92  & 96  & 80 & $\mathbf{89.3 \pm 3.6}$ \\
      & Verification accuracy & 96 & 92 & 88 & $\mathbf{92.0 \pm 3.1}$ \\
      & Reset SR              & 84  & 84  & 56 & $\mathbf{74.7 \pm 5.0}$ \\
    \bottomrule
  \end{tabular}
\end{table}

\begin{figure*}[!t]
  \centering
  \includegraphics[width=\textwidth]{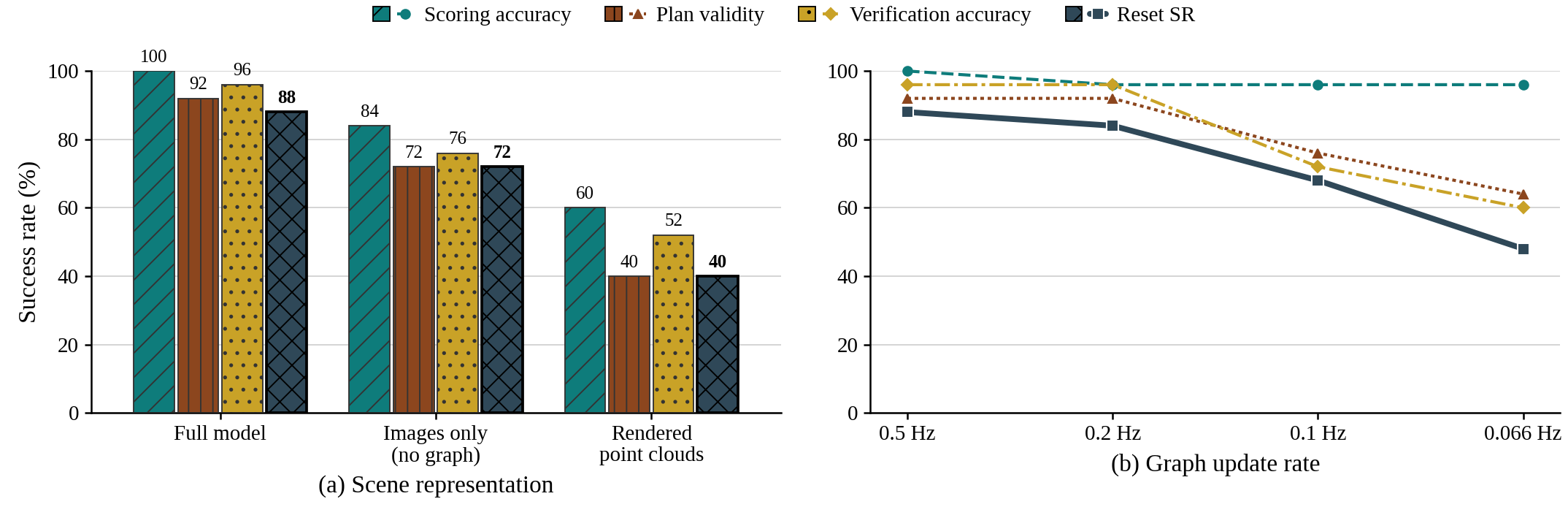}
  \caption{Ablation of the scene graph on Drawer-Store, 25 episodes per
  configuration. (a) Replacing the graph with the images themselves, or with
  rendered point clouds, lowers every metric. (b) As the graph update rate
  drops, plan validity and reset SR fall while scoring accuracy stays flat.}
  \label{fig:ablation}
\end{figure*}

\subsection{Composing Skills for Held-Out Tasks (Q2)}
\label{sec:q2}

Table~\ref{tab:compositional} reports three held-out tasks that
\sysname has not previously evaluated. These tasks retain the same objects but
change the task goals. In the drawer scene, a successful rollout leaves either
the lemon on the table with the drawer closed or the strawberry on the table
rather than on the plate. In the kitchen scene, it leaves the stock pot on the
stove with its lid on top without carrot. These task goals produce terminal states that require
reset sequences different from those of the original tasks. \sysname reaches
74.7\% reset SR and 89.3\% plan validity on them without a single new
demonstration. AutoEval reaches 1.3\%, because it fits one reset policy to
the skill sequence of the task it was trained on, and that sequence no longer
applies. The difference lies in what each system learns: a library of atomic
skills recombines into a new order, while a single policy cannot. This is the
compositional generalization that the hierarchy is meant to provide.

The same property also changes what a demonstration has to contain. A demonstration of
one atomic skill is far shorter than a demonstration of a whole reset, so
resetting the four original tasks costs \sysname 105k timesteps against 138k for
AutoEval, 24\% less data, and the same skills serve every task of a scene.

\subsection{Ablation of the Scene Graph (Q3)}
\label{sec:q3}

Fig.~\ref{fig:ablation} ablates the scene graph on Drawer-Store. Replacing the
graph with the images themselves lowers plan validity from 92\% to 72\%,
verification accuracy from 96\% to 76\%, and reset SR from 88\% to 72\%.
Passing rendered point clouds instead lowers them further, to 40\%, 52\%, and
40\%, so the gain does not come from three-dimensional appearance but from
turning geometry into explicit relations.

The rate at which \sysname updates the graph matters as well. Performance holds
at 0.2\,Hz, and below that rate plan validity falls from 92\% to 64\%,
verification accuracy from 96\% to 60\%, and reset SR from 88\% to 48\%, while
scoring accuracy stays at 96\%. This follows
Section~\ref{sec:scene_graph}: a plan needs the states that the rollout passed
through, whereas the completion score can often be recovered from the last few
keyframes.

%% file: section/05_conclusion.tex
\section{Conclusion}
\label{sec:conclusion}

We presented \sysname, an evaluation harness that scores a long-horizon rollout
on real hardware and restores the scene for the next one with little human
involvement.
\sysname builds a spatial scene graph online from RGB-D keyframes, and an LLM
reads that graph to score the rollout, to plan a reset over a library of atomic
skills, and to verify that the reset succeeded. Planning over a library makes
the demonstration cost scale with the size of that library rather than with the
number of states a rollout can terminate in, and on four real tasks \sysname
restores the scene more reliably than a per-task reset policy while cutting
the operator time by 72\%. On three held-out tasks, it recombines the same skills
without collecting anything new, and an ablation attributes the gain to the
graph, since replacing it with images or with rendered point clouds lowers every
metric. Two limitations remain. A reset that needs a skill the rollout never performs
still calls for demonstrations of its own, and \sysname still needs a handful of
reference cases written for each task, although it trains no classifier and
collects no labeled images.
Reversing the forward demonstrations that a task already provides is a
promising way to obtain the missing skills without a new collection effort.

%% file: main.bib
@article{intelligence2026pi,
  title={{$\pi_{0.7}$}: A Steerable Generalist Robotic Foundation Model with Emergent Capabilities},
  author={{Physical Intelligence} and Ai, Bo and Amin, Ali and Aniceto, Raichelle and Balakrishna, Ashwin and Balke, Greg and Black, Kevin and Bokinsky, George and Cao, Shihao and Charbonnier, Thomas and others},
  journal={arXiv preprint arXiv:2604.15483},
  year={2026}
}

@inproceedings{kim2025fine,
  title={Fine-Tuning Vision-Language-Action Models: Optimizing Speed and Success},
  author={Kim, Moo Jin and Finn, Chelsea and Liang, Percy},
  booktitle={Robotics: Science and Systems XXI},
  year={2025},
  doi={10.15607/RSS.2025.XXI.017}
}

@article{kim2026cosmos,
  title={{Cosmos Policy}: Fine-Tuning Video Models for Visuomotor Control and Planning},
  author={Kim, Moo Jin and Gao, Yihuai and Lin, Tsung-Yi and Lin, Yen-Chen and Ge, Yunhao and Lam, Grace and Liang, Percy and Song, Shuran and Liu, Ming-Yu and Finn, Chelsea and Gu, Jinwei},
  journal={arXiv preprint arXiv:2601.16163},
  year={2026}
}

@article{zhang2026native,
  title={Native video-action pretraining for generalizable robot control},
  author={Zhang, Qihang and Li, Lin and Zhang, Luyao and Yang, Shuai and Luo, Yiming and Li, Shuaiting and Wang, Ruilin and Wang, Junke and Shao, Jiahao and Xu, Gangwei and others},
  journal={arXiv preprint arXiv:2607.08639},
  year={2026}
}

@article{jangir2025robotarena,
  title={{RobotArena} $\infty$: Scalable Robot Benchmarking via Real-to-Sim Translation},
  author={Jangir, Yash and Zhang, Yidi and Lo, Pang-Chi and Yamazaki, Kashu and Zhang, Chenyu and Tu, Kuan-Hsun and Ke, Tsung-Wei and Ke, Lei and Bisk, Yonatan and Fragkiadaki, Katerina},
  journal={arXiv preprint arXiv:2510.23571},
  year={2025}
}

@article{chen2026robodojo,
  title={{RoboDojo}: A Unified Sim-and-Real Benchmark for Comprehensive Evaluation of Generalist Robot Manipulation Policies},
  author={Chen, Tianxing and Chen, Yue and Li, Zixuan and Tang, Junyuan and Su, Kailun and Lu, Haoran and Wan, Weijie and Chen, Baijun and Liu, Songling and Yan, Haowen and others},
  journal={arXiv preprint arXiv:2607.04434},
  year={2026}
}

@article{mittal2025isaac,
  title={{Isaac Lab}: A {GPU}-Accelerated Simulation Framework for Multi-Modal Robot Learning},
  author={Mittal, Mayank and Roth, Pascal and Tigue, James and Richard, Antoine and Zhang, Octi and Du, Peter and Serrano-Munoz, Antonio and Yao, Xinjie and Zurbr{\"u}gg, Ren{\'e} and Rudin, Nikita and others},
  journal={arXiv preprint arXiv:2511.04831},
  year={2025}
}

@article{tao2024maniskill3,
  title={{ManiSkill3}: {GPU} Parallelized Robotics Simulation and Rendering for Generalizable Embodied {AI}},
  author={Tao, Stone and Xiang, Fanbo and Shukla, Arth and Qin, Yuzhe and Hinrichsen, Xander and Yuan, Xiaodi and Bao, Chen and Lin, Xinsong and Liu, Yulin and Chan, Tse-kai and others},
  journal={arXiv preprint arXiv:2410.00425},
  year={2024}
}

@article{zhu2020robosuite,
  title={{robosuite}: A Modular Simulation Framework and Benchmark for Robot Learning},
  author={Zhu, Yuke and Wong, Josiah and Mandlekar, Ajay and Mart{\'\i}n-Mart{\'\i}n, Roberto and Joshi, Abhishek and Lin, Kevin and Maddukuri, Abhiram and Nasiriany, Soroush and Zhu, Yifeng},
  journal={arXiv preprint arXiv:2009.12293},
  year={2020}
}

@inproceedings{zhou2025autoeval,
  title={{AutoEval}: Autonomous Evaluation of Generalist Robot Manipulation Policies in the Real World},
  author={Zhou, Zhiyuan and Atreya, Pranav and Tan, You Liang and Pertsch, Karl and Levine, Sergey},
  booktitle={Proceedings of The 9th Conference on Robot Learning},
  pages={1997--2017},
  year={2025},
  volume={305},
  series={Proceedings of Machine Learning Research},
  publisher={PMLR}
}

@article{barreiros2026careful,
  title={A careful examination of large behavior models for multitask dexterous manipulation},
  author={Barreiros, Jose and Beaulieu, Andrew and Bhat, Aditya and Cory, Rick and Cousineau, Eric and Dai, Hongkai and Fang, Ching-Hsin and Hashimoto, Kunimatsu and Irshad, Muhammad Zubair and Itkina, Masha and others},
  journal={Science Robotics},
  volume={11},
  number={113},
  pages={eaea6201},
  year={2026},
  publisher={American Association for the Advancement of Science}
}

@inproceedings{khargonkar2024scenereplica,
  title={{SceneReplica}: Benchmarking Real-World Robot Manipulation by Creating Replicable Scenes},
  author={Khargonkar, Ninad and Allu, Sai Haneesh and Lu, Yangxiao and P, Jishnu Jaykumar and Prabhakaran, Balakrishnan and Xiang, Yu},
  booktitle={2024 IEEE International Conference on Robotics and Automation (ICRA)},
  pages={8258--8264},
  year={2024},
  organization={IEEE}
}

@article{yang2026lilo,
  title={{LiLo-VLA}: Compositional Long-Horizon Manipulation via Linked Object-Centric Policies},
  author={Yang, Yue and Cheng, Shuo and Fang, Yu and Bharadhwaj, Homanga and Ding, Mingyu and Bertasius, Gedas and Szafir, Daniel},
  journal={arXiv preprint arXiv:2602.21531},
  year={2026}
}

@article{grislain2025failsense,
  title={{I-FailSense}: Towards General Robotic Failure Detection with Vision-Language Models},
  author={Grislain, Cl{\'e}mence and Rahimi, Hamed and Sigaud, Olivier and Chetouani, Mohamed},
  journal={arXiv preprint arXiv:2509.16072},
  year={2025}
}

@inproceedings{chen2024spatialvlm,
  title={{SpatialVLM}: Endowing Vision-Language Models with Spatial Reasoning Capabilities},
  author={Chen, Boyuan and Xu, Zhuo and Kirmani, Sean and Ichter, Brian and Sadigh, Dorsa and Guibas, Leonidas and Xia, Fei},
  booktitle={2024 IEEE/CVF Conference on Computer Vision and Pattern Recognition (CVPR)},
  pages={14455--14465},
  year={2024}
}

@inproceedings{pothiraj2025capture,
  title={{CAPTURe}: Evaluating Spatial Reasoning in Vision Language Models via Occluded Object Counting},
  author={Pothiraj, Atin and Stengel-Eskin, Elias and Cho, Jaemin and Bansal, Mohit},
  booktitle={2025 IEEE/CVF International Conference on Computer Vision (ICCV)},
  pages={8001--8010},
  year={2025},
  organization={IEEE}
}

@inproceedings{ravi2025sam,
  title={{SAM} 2: Segment Anything in Images and Videos},
  author={Ravi, Nikhila and Gabeur, Valentin and Hu, Yuan-Ting and Hu, Ronghang and Ryali, Chaitanya and Ma, Tengyu and Khedr, Haitham and R{\"a}dle, Roman and Rolland, Chloe and Gustafson, Laura and others},
  booktitle={International Conference on Learning Representations},
  volume={2025},
  pages={28085--28128},
  year={2025}
}

@article{oquab2023dinov2,
  title={{DINOv2}: Learning Robust Visual Features without Supervision},
  author={Oquab, Maxime and Darcet, Timoth{\'e}e and Moutakanni, Th{\'e}o and Vo, Huy and Szafraniec, Marc and Khalidov, Vasil and Fernandez, Pierre and Haziza, Daniel and Massa, Francisco and El-Nouby, Alaaeldin and others},
  journal={Transactions on Machine Learning Research},
  volume={2024},
  year={2024}
}

@article{garrett2021integrated,
  title={Integrated task and motion planning},
  author={Garrett, Caelan Reed and Chitnis, Rohan and Holladay, Rachel and Kim, Beomjoon and Silver, Tom and Kaelbling, Leslie Pack and Lozano-P{\'e}rez, Tom{\'a}s},
  journal={Annual Review of Control, Robotics, and Autonomous Systems},
  volume={4},
  number={1},
  pages={265--293},
  year={2021},
  publisher={Annual Reviews}
}

@inproceedings{kaelbling2011hierarchical,
  title={Hierarchical task and motion planning in the now},
  author={Kaelbling, Leslie Pack and Lozano-P{\'e}rez, Tom{\'a}s},
  booktitle={2011 IEEE international conference on robotics and automation},
  pages={1470--1477},
  year={2011},
  organization={IEEE}
}

@inproceedings{dalal2024plan,
  title={{Plan-Seq-Learn}: Language Model Guided RL for Solving Long Horizon Robotics Tasks},
  author={Dalal, Murtaza and Chiruvolu, Tarun and Chaplot, Devendra Singh and Salakhutdinov, Ruslan},
  booktitle={International Conference on Learning Representations (ICLR)},
  year={2024}
}

@inproceedings{dalal2025local,
  title={Local Policies Enable Zero-Shot Long-Horizon Manipulation},
  author={Dalal, Murtaza and Liu, Min and Talbott, Walter and Chen, Chen and Pathak, Deepak and Zhang, Jian and Salakhutdinov, Ruslan},
  booktitle={2025 IEEE International Conference on Robotics and Automation (ICRA)},
  pages={13875--13882},
  year={2025},
  organization={IEEE}
}

@article{cheng2023league,
  title={{LEAGUE}: Guided Skill Learning and Abstraction for Long-Horizon Manipulation},
  author={Cheng, Shuo and Xu, Danfei},
  journal={IEEE Robotics and Automation Letters},
  volume={8},
  number={10},
  pages={6451--6458},
  year={2023},
  publisher={IEEE}
}

@inproceedings{mishra2023generative,
  title={Generative Skill Chaining: Long-Horizon Skill Planning with Diffusion Models},
  author={Mishra, Utkarsh Aashu and Xue, Shangjie and Chen, Yongxin and Xu, Danfei},
  booktitle={Conference on Robot Learning},
  pages={2905--2925},
  year={2023},
  organization={PMLR}
}

@inproceedings{fan2025long,
  title={{Long-VLA}: Unleashing Long-Horizon Capability of Vision Language Action Model for Robot Manipulation},
  author={Fan, Yiguo and Bai, Shuanghao and Tong, Xinyang and Ding, Pengxiang and Zhu, Yuyang and Lu, Hongchao and Dai, Fengqi and Zhao, Wei and Liu, Yang and Huang, Siteng and Fan, Zhaoxin and Chen, Badong and Wang, Donglin},
  booktitle={Proceedings of The 9th Conference on Robot Learning},
  pages={2018--2037},
  year={2025},
  organization={PMLR}
}

@article{yang2025boss,
  title={{BOSS}: Benchmark for Observation Space Shift in Long-Horizon Task},
  author={Yang, Yue and Zhao, Linfeng and Ding, Mingyu and Bertasius, Gedas and Szafir, Daniel},
  journal={IEEE Robotics and Automation Letters},
  volume={10},
  number={9},
  pages={8882--8889},
  year={2025},
  publisher={IEEE}
}

@inproceedings{liu2023reflect,
  title={{REFLECT}: Summarizing Robot Experiences for Failure Explanation and Correction},
  author={Liu, Zeyi and Bahety, Arpit and Song, Shuran},
  booktitle={Conference on Robot Learning (CoRL)},
  pages={3468--3484},
  year={2023}
}

@misc{atreya2025roboarenadistributedrealworldevaluation,
      title={{RoboArena}: Distributed Real-World Evaluation of Generalist Robot Policies}, 
      author={Pranav Atreya and Karl Pertsch and Tony Lee and Moo Jin Kim and Arhan Jain and Artur Kuramshin and Clemens Eppner and Cyrus Neary and Edward Hu and Fabio Ramos and Jonathan Tremblay and Kanav Arora and Kirsty Ellis and Luca Macesanu and Marcel Torne Villasevil and Matthew Leonard and Meedeum Cho and Ozgur Aslan and Shivin Dass and Jie Wang and William Reger and Xingfang Yuan and Xuning Yang and Abhishek Gupta and Dinesh Jayaraman and Glen Berseth and Kostas Daniilidis and Roberto Martin-Martin and Youngwoon Lee and Percy Liang and Chelsea Finn and Sergey Levine},
      year={2025},
      eprint={2506.18123},
      archivePrefix={arXiv},
      primaryClass={cs.RO},
      url={https://arxiv.org/abs/2506.18123}, 
}

@inproceedings{heo2023furniturebenchreproduciblerealworldbenchmark,
      title={{FurnitureBench}: Reproducible Real-World Benchmark for Long-Horizon Complex Manipulation}, 
      author={Minho Heo and Youngwoon Lee and Doohyun Lee and Joseph J. Lim},
      booktitle={Robotics: Science and Systems XIX},
      year={2023},
      doi={10.15607/RSS.2023.XIX.041},
}

@misc{yang2019replabreproduciblelowcostarm,
      title={{REPLAB}: A Reproducible Low-Cost Arm Benchmark Platform for Robotic Learning}, 
      author={Brian Yang and Jesse Zhang and Vitchyr Pong and Sergey Levine and Dinesh Jayaraman},
      year={2019},
      eprint={1905.07447},
      archivePrefix={arXiv},
      primaryClass={cs.RO},
      url={https://arxiv.org/abs/1905.07447}, 
}

@misc{xiao2026enpireagenticrobotpolicy,
      title={{ENPIRE}: Agentic Robot Policy Self-Improvement in the Real World}, 
      author={Wenli Xiao and Jia Xie and Tonghe Zhang and Haotian Lin and Letian "Max" Fu and Haoru Xue and Jalen Lu and Yi Yang and Cunxi Dai and Zi Wang and Jimmy Wu and Guanzhi Wang and S. Shankar Sastry and Ken Goldberg and Linxi "Jim" Fan and Yuke Zhu and Guanya Shi},
      year={2026},
      eprint={2606.19980},
      archivePrefix={arXiv},
      primaryClass={cs.AI},
      url={https://arxiv.org/abs/2606.19980}, 
}

@misc{wang2026radarclosedlooproboticdata,
      title={{RADAR}: Closed-Loop Robotic Data Generation via Semantic Planning and Autonomous Causal Environment Reset}, 
      author={Yongzhong Wang and Keyu Zhu and Yong Zhong and Liqiong Wang and Jinyu Yang and Feng Zheng},
      year={2026},
      eprint={2603.11811},
      archivePrefix={arXiv},
      primaryClass={cs.RO},
      url={https://arxiv.org/abs/2603.11811}, 
}

@misc{wang2026zero2skillbootstrappingrobotskills,
      title={{Zero2Skill}: Bootstrapping Robot Skills through Autonomous Data Collection, Training, and Deployment}, 
      author={Boyuan Wang and Zhenyuan Zhang and Zhiqin Yang and Peijun Gu and Shuya Wang and Xiaofeng Wang and Xianghui Ze and Yifan Chang and Guosheng Zhao and Jiangnan Shao and Guan Huang and Hengyu Liu and Yonggang Zhang and Wei Xue and Chunyuan Guan and Chenglin Pu and Yike Guo and Xingang Wang and Zheng Zhu},
      year={2026},
      eprint={2607.14047},
      archivePrefix={arXiv},
      primaryClass={cs.RO},
      url={https://arxiv.org/abs/2607.14047}, 
}

@misc{zhang2026harnessvlasteeringfrozen,
      title={{Harness VLA}: Steering Frozen VLAs into Reliable Manipulation Primitives via Memory-Guided Agents}, 
      author={Yixian Zhang and Huanming Zhang and Feng Gao and Xiao Li and Zhihao Liu and Chunyang Zhu and Jiaxing Qiu and Yuchen Yan and Jiyuan Liu and Wenhao Tang and Zhengru Fang and Yi Nie and Changxu Wei and Yu Wang and Wenbo Ding and Chao Yu},
      year={2026},
      eprint={2607.08448},
      archivePrefix={arXiv},
      primaryClass={cs.RO},
      url={https://arxiv.org/abs/2607.08448}, 
}

@misc{liu2026guavaeffectiveuniversalharness,
      title={{Guava}: An Effective and Universal Harness for Embodied Manipulation}, 
      author={Haowen Liu and Xirui Li and Shaoxiong Yao and Peng Shi and Tianyi Zhou and Jia-Bin Huang and Furong Huang and Jiayuan Mao},
      year={2026},
      eprint={2606.18363},
      archivePrefix={arXiv},
      primaryClass={cs.RO},
      url={https://arxiv.org/abs/2606.18363}, 
}

@misc{jiang2026robotttcontextscalingrobot,
      title={{RoboTTT}: Context Scaling for Robot Policies}, 
      author={Yunfan Jiang and Yevgen Chebotar and Ruijie Zheng and Fengyuan Hu and Yunhao Ge and Jimmy Wu and Tianyuan Dai and Scott Reed and Li Fei-Fei and Yuke Zhu and Linxi "Jim" Fan},
      year={2026},
      eprint={2607.15275},
      archivePrefix={arXiv},
      primaryClass={cs.RO},
      url={https://arxiv.org/abs/2607.15275}, 
}

@misc{zhang2026foresightfailuredetectionlonghorizon,
      title={{Foresight}: Failure Detection for Long-Horizon Robotic Manipulation with Action-Conditioned World Model Latents}, 
      author={Haoran Zhang and Yifu Lu and Boyang Wang and Xuhui Kang and Yen-Ling Kuo and Zezhou Cheng and Mengdi Wang and Odest Chadwicke Jenkins},
      year={2026},
      eprint={2606.23085},
      archivePrefix={arXiv},
      primaryClass={cs.RO},
      url={https://arxiv.org/abs/2606.23085}, 
}

@inproceedings{cheng2024nodtampgeneralizablelonghorizonplanning,
      title={{NOD-TAMP}: Generalizable Long-Horizon Planning with Neural Object Descriptors}, 
      author={Shuo Cheng and Caelan Reed Garrett and Ajay Mandlekar and Danfei Xu},
      booktitle={Conference on Robot Learning (CoRL)},
      series={Proceedings of Machine Learning Research},
      volume={270},
      pages={1310--1339},
      publisher={PMLR},
      year={2024},
}

@article{yu2025scene,
  title={Scene graph-guided proactive replanning for failure-resilient embodied agent},
  author={Yu, Che Rin and Chae, Daewon and Seo, Dabin and Lee, Sangwon and Im, Hyeongwoo and Kim, Jinkyu},
  journal={arXiv preprint arXiv:2508.11286},
  year={2025}
}

@inproceedings{ju2026momagraph,
  title={MomaGraph: State-Aware Unified Scene Graphs with Vision-Language Models for Embodied Task Planning},
  author={Ju, Yuanchen and Liang, Yongyuan and Wang, Yen-Jen and Nandiraju, Gireesh and Ju, Yuanliang and Lee, Seungjae Jay and Gu, Qiao and Hsieh, Elvis and Huang, Furong and Sreenath, Koushil},
  booktitle={International Conference on Learning Representations},
  volume={2026},
  pages={96361--96386},
  year={2026}
}

@inproceedings{zhu2021hierarchical,
  title={Hierarchical planning for long-horizon manipulation with geometric and symbolic scene graphs},
  author={Zhu, Yifeng and Tremblay, Jonathan and Birchfield, Stan and Zhu, Yuke},
  booktitle={2021 IEEE International Conference on Robotics and Automation (ICRA)},
  pages={6541--6548},
  year={2021},
  organization={Ieee}
}

@article{ren2024grounded,
  title={Grounded sam: Assembling open-world models for diverse visual tasks},
  author={Ren, Tianhe and Liu, Shilong and Zeng, Ailing and Lin, Jing and Li, Kunchang and Cao, He and Chen, Jiayu and Huang, Xinyu and Chen, Yukang and Yan, Feng and others},
  journal={arXiv preprint arXiv:2401.14159},
  year={2024}
}

@inproceedings{liu2024grounding,
  title={Grounding dino: Marrying dino with grounded pre-training for open-set object detection},
  author={Liu, Shilong and Zeng, Zhaoyang and Ren, Tianhe and Li, Feng and Zhang, Hao and Yang, Jie and Jiang, Qing and Li, Chunyuan and Yang, Jianwei and Su, Hang and others},
  booktitle={European conference on computer vision},
  pages={38--55},
  year={2024},
  organization={Springer}
}

@article{reuss2023goal,
  title={Goal-conditioned imitation learning using score-based diffusion policies},
  author={Reuss, Moritz and Li, Maximilian and Jia, Xiaogang and Lioutikov, Rudolf},
  journal={arXiv preprint arXiv:2304.02532},
  year={2023}
}

@article{zheng2025xvla,
  title={X-VLA: Soft-Prompted Transformer as Scalable Cross-Embodiment Vision-Language-Action Model},
  author={Zheng, Jinliang and Li, Jianxiong and Wang, Zhihao and Liu, Dongxiu and Kang, Xirui and Feng, Yuchun and Zheng, Yinan and Zou, Jiayin and Chen, Yilun and Zeng, Jia and Zhang, Ya-Qin and Pang, Jiangmiao and Liu, Jingjing and Wang, Tai and Zhan, Xianyuan},
  journal={arXiv preprint arXiv:2510.10274},
  year={2025}
}
